\documentclass[conference]{IEEEtran}
\IEEEoverridecommandlockouts 

\usepackage{cite} 
\usepackage{amsmath,amssymb,amsfonts} 
\usepackage{graphicx} 
\usepackage{textcomp} 
\usepackage{xcolor} 
\usepackage{url} 
\usepackage{booktabs} 
\usepackage{multirow} 

\usepackage{microtype}

\usepackage{hyperref} 

\begin{document}

\title{SI-Agent: An Agentic Framework for Feedback-Driven Generation and Tuning of Human-Readable System Instructions for Large Language Models
}

\makeatletter
\newcommand{\linebreakand}{%
  \end{@IEEEauthorhalign}
  \hfill\mbox{}\par
  \mbox{}\hfill\begin{@IEEEauthorhalign}
}
\makeatother

\author{
  \IEEEauthorblockN{Jeshwanth Challagundla}
  \IEEEauthorblockA{\textit{University of Texas at Arlington}\\ 
                   TX, USA \\
                   jeshwanth.challagundla@mavs.uta.edu} 
  \and
  \IEEEauthorblockN{Mantek Singh}
  \IEEEauthorblockA{\textit{Liverpool John Moores University}\\
                   Liverpool, England \\
                   mantek.singh2@gmail.com}
  \and
  \IEEEauthorblockN{Siddharth Raina}
  \IEEEauthorblockA{\textit{Carnegie Mellon University}\\
                   PA, USA \\
                   sraina1@alumni.cmu.edu}
  \linebreakand
  \IEEEauthorblockN{Smarth Behl}
  \IEEEauthorblockA{\textit{Bits Pilani}\\
                   CA, USA \\
                   smarthbehl@gmail.com}
\and
 
  \IEEEauthorblockN{FNU Harsh}
  \IEEEauthorblockA{WA, USA \\
                   hars@outlook.com}
 \and
  \IEEEauthorblockN{Jasmin Jarsania}
  \IEEEauthorblockA{CA, USA \\
                   jasmin.jarsania@gmail.com}
}

\maketitle

\begin{abstract}
System Instructions (SIs), or system prompts, are pivotal for guiding Large Language Models (LLMs) but manual crafting is resource-intensive and often suboptimal. Existing automated methods frequently generate non-human-readable "soft prompts," sacrificing interpretability. This paper introduces SI-Agent, a novel agentic framework designed to automatically generate and iteratively refine \emph{human-readable} SIs through a feedback-driven loop. SI-Agent employs three collaborating agents: an Instructor Agent, an Instruction Follower Agent (target LLM), and a Feedback/Reward Agent evaluating task performance and optionally SI readability. The framework utilizes iterative cycles where feedback guides the Instructor's refinement strategy (e.g., LLM-based editing, evolutionary algorithms). We detail the framework's architecture, agent roles, the iterative refinement process, and contrast it with existing methods. We present experimental results validating SI-Agent's effectiveness, focusing on metrics for task performance, SI readability, and efficiency. Our findings indicate that SI-Agent generates effective, readable SIs, offering a favorable trade-off between performance and interpretability compared to baselines. Potential implications include democratizing LLM customization and enhancing model transparency. Challenges related to computational cost and feedback reliability are acknowledged.
\end{abstract}


%


\section{Introduction}

\subsection{The Crucial Role and Challenges of System Instructions (SIs)}

Large Language Models (LLMs) have demonstrated remarkable capabilities across a vast spectrum of natural language tasks. A key factor influencing their performance and behavior is the set of instructions provided alongside the primary task input, commonly referred to as System Instructions (SIs), system prompts, or meta-prompts \cite{ref1, ref2}. These instructions serve as a powerful mechanism to guide the model's persona, define its objectives, specify constraints, delineate desired output formats, and provide crucial context \cite{ref4, ref5}. Effective SIs can significantly enhance task performance, steer the model towards desired behaviors like truthfulness or specific reasoning patterns (e.g., Chain-of-Thought), and adapt pre-trained models to downstream tasks without necessitating modifications to the underlying model parameters \cite{ref1, ref2, ref7}.

Despite their importance, the process of creating optimal SIs, often termed prompt engineering, presents a significant bottleneck in the practical deployment of LLMs \cite{ref7, ref8, ref9}. Manual prompt engineering is an iterative, often heuristic-driven process that demands considerable time, human effort, and specialized expertise \cite{ref4, ref7}. Crafting effective prompts requires not only domain knowledge related to the task but also a nuanced understanding of the specific LLM's behavior and sensitivities \cite{ref7}. The process frequently involves extensive trial-and-error, leading to prompts that can become overly complex, convoluted, or "messy," necessitating careful pruning and ongoing maintenance as tasks or models evolve \cite{ref4, ref5, ref9}. Achieving consistent, high performance across diverse inputs or different LLM architectures through manual tuning remains a substantial challenge \cite{ref10}. While guidelines exist \cite{ref4, ref5}, translating these principles into optimal textual instructions manually is difficult to scale effectively and systematically explore the vast instruction space \cite{ref10}.

\subsection{The Need for Automated, Human-Readable SI Generation}

The inherent limitations of manual SI crafting necessitate the development of automated approaches \cite{ref7, ref8, ref9, ref10, ref18}. Automation promises to alleviate the burden on human engineers, offering potential benefits in terms of efficiency, consistency, and the ability to discover superior SIs \cite{ref10, ref13, ref14, ref18}.

However, a critical distinction exists within automated methods regarding the nature of the optimized prompt. One prominent line of research optimizes "soft prompts" or "continuous prompts" \cite{ref19, ref21, ref24}. Techniques like Prompt Tuning \cite{ref21, ref24} and Prefix-Tuning \cite{ref19, ref20} learn non-human-readable continuous vector representations. While parameter-efficient and performant \cite{ref24}, this lack of interpretability hinders debugging, trust, and verification.

Conversely, there is a compelling need for automated methods that generate and optimize \emph{human-readable} SIs composed of discrete natural language text \cite{ref29, ref32}. Readability is paramount for understanding model behavior, debugging, auditing, and collaborative development. While some techniques produce discrete prompts \cite{ref13, ref14, ref29, ref32}, a dedicated framework explicitly prioritizing human readability alongside performance, using structured interactions, remains needed. This tension between peak performance via potentially opaque methods and human understanding via readable instructions motivates our work.

\subsection{Proposed Solution: An Agentic Framework (SI-Agent)}

To address the need for automated generation and tuning of effective yet understandable SIs, this paper introduces SI-Agent, a novel agentic framework. SI-Agent leverages principles from multi-agent systems (MAS) \cite{ref35, ref60} to structure the complex optimization process, employing specialized agents collaborating within an iterative feedback loop.

The SI-Agent framework comprises three core components (illustrated conceptually in Figure \ref{fig:architecture}): 

\textbf{1) Instructor Agent:} Responsible for generating and iteratively refining human-readable SIs based on feedback.

\textbf{2) Instruction Follower Agent:} The target LLM executing tasks using the SI, providing a performance signal.

\textbf{3) Feedback/Reward Agent:} Evaluates task performance and optionally SI readability, generating feedback for the Instructor.

This multi-agent structure enables a modular approach, separating SI generation, execution, and evaluation \cite{ref13, ref14}. Inspired by MAS concepts \cite{ref35} and frameworks like AutoGen \cite{ref36}, it uses specialized roles and structured communication. The feedback signal guides the Instructor Agent's refinement process towards optimal performance and readability.

\begin{figure*}[tbp]
\centering
\includegraphics[width=1\textwidth]{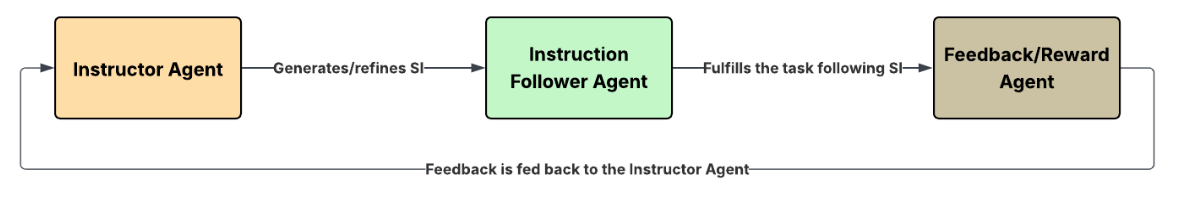}
\caption{Architecture of the SI-Agent Framework. The Instructor Agent generates SI$_i$, the Follower Agent executes Task$_{input}$ using SI$_i$ producing Output$_i$, and the Feedback Agent evaluates Output$_i$ (and optionally SI$_i$) to generate Feedback$_i$, which guides the Instructor's refinement for SI$_{i+1}$.}
\label{fig:architecture}
\end{figure*}

\subsection{Contributions}
The primary contributions of this paper are:

\textbf{1) Novel Agentic Framework:} We introduce SI-Agent, a dedicated multi-agent framework specifically designed for the automatic generation and optimization of human-readable System Instructions for LLMs.

\textbf{2) Feedback-Driven Iterative Refinement:} We detail the collaborative process where structured feedback on task performance and SI quality drives the iterative refinement of human-readable SIs, explicitly aiming to balance effectiveness and interpretability.

\textbf{3) Comparative Analysis:} We provide a structured comparison of the proposed framework with existing manual and automated approaches.

\textbf{4) Experimental Validation:} We present an empirical evaluation demonstrating the framework's effectiveness, efficiency, and the quality of the generated SIs across a diverse set of benchmark tasks.

This work aims to bridge the gap between automated LLM customization and interpretable control, offering a principled approach to generating effective and readable SIs, supported by experimental validation.

\section{Related Work}
\subsection{Manual System Instruction / Prompt Engineering}

Manual prompt engineering remains a common practice for controlling LLMs. It involves crafting natural language instructions to elicit desired responses. Standard techniques include:

\textbf{Zero-Shot Prompting:} Providing the task instruction directly without examples \cite{ref2, ref4}.

\textbf{Few-Shot Prompting:} Including a small number of input-output examples (demonstrations) within the prompt to guide the model's behavior and specify the desired format or style \cite{ref2, ref4, ref5}. While effective, this increases context length \cite{ref2}.

\textbf{Chain-of-Thought (CoT) Prompting:} Encouraging the model to generate intermediate reasoning steps before arriving at the final answer, often improving performance on complex reasoning tasks \cite{ref1, ref2, ref4, ref5, ref40}. This can be triggered by simple phrases like "Let's think step by step" \cite{ref13, ref40}.

Effective manual prompts often require clear goal definition, sufficient context, structured instructions (breaking down tasks), role-playing, and constraints (e.g., length limits, format specifications) \cite{ref4, ref5}. However, this process is fundamentally limited by human intuition, effort, and the difficulty of exploring the vast prompt space \cite{ref7, ref10}. It often requires extensive iteration and testing \cite{ref4, ref5}, and prompts may lack robustness across different inputs or models \cite{ref7, ref9, ref10}.

\subsection{Automated Prompt Optimization (Human-Readable / Discrete)}

To overcome the limitations of manual engineering, various methods have been proposed to automatically generate or optimize discrete, human-readable text prompts. These methods typically treat the target LLM as a black-box, interacting with it via its input-output interface without requiring access to gradients or internal states \cite{ref42, ref43}. This black-box nature makes them broadly applicable, especially to proprietary API-based models.

\textbf{LLM-Based Generation/Refinement:} These methods leverage the generative capabilities of LLMs themselves to create or improve prompts, mirroring their use in other complex generation tasks such as synthetic data creation \cite{singh2024llms}. Examples specific to prompt engineering include: 

\textbf{APE (Automatic Prompt Engineer):} Frames instruction generation as a black-box optimization problem \cite{ref13}. It uses an "inference" LLM to propose instruction candidates based on input-output demonstrations of the target task \cite{ref13, ref16, ref45}. These candidates are then evaluated by executing them with a "target" LLM on a subset of data, and the best-performing instruction is selected based on a score function, such as execution accuracy (0-1 loss) or log probability \cite{ref13}. APE has been shown to discover prompts outperforming human-engineered ones, including effective zero-shot CoT prompts \cite{ref13}. The process often involves templates for generation and evaluation.

\textbf{OPRO (Optimization by PROmpting):} Uses an LLM \emph{as} the optimizer \cite{ref14}. It feeds the optimizer LLM a "meta-prompt" containing a natural language description of the optimization goal (e.g., maximize accuracy), the history of previously generated prompts and their scores (optimization trajectory), and task examples \cite{ref14}. The LLM leverages this context to generate new, potentially improved prompt candidates in an iterative loop \cite{ref14}. OPRO demonstrated strong performance on reasoning benchmarks \cite{ref14}.

\textbf{PE2:} Focuses on improving the LLM's reasoning ability during prompt optimization by enriching the meta-prompt with detailed task descriptions, context specifications, and reasoning templates \cite{ref8}.

\textbf{STRAGO:} Employs a reflection-based approach where an LLM analyzes successful and failed examples from a current prompt to generate "experiences" and actionable strategies for guiding the next prompt refinement step, aiming to prevent performance degradation on previously successful cases ("prompt drifting") \cite{ref42}.

\textbf{Evolutionary Algorithms:} These methods apply principles of biological evolution to search the prompt space. Examples include:

\textbf{Promptbreeder:} Evolves a population of "task-prompts" over generations \cite{ref10}. It uses an LLM to apply various mutation operators (e.g., paraphrasing, combining prompts, generating variations based on lineage). Crucially, the mutation process itself is guided by "mutation-prompts" which are also co-evolved in a self-referential manner, allowing the system to learn \emph{how} to mutate prompts effectively \cite{ref10}. Fitness is evaluated based on task accuracy on a training batch. Promptbreeder has shown strong results on reasoning benchmarks and can be applied to multimodal tasks \cite{ref10, ref41}.

\textbf{EvoPrompt:} Demonstrates that evolutionary algorithms can systematically evolve discrete prompts to improve task accuracy \cite{ref41, ref42}.

\textbf{APEX:} Specifically targets the optimization of \emph{long} prompts (containing instructions, demonstrations, and CoT) by iteratively replacing sentences with semantically equivalent alternatives generated by an LLM, using beam search to manage the process \cite{ref50}.

\textbf{Reinforcement Learning:} These methods frame prompt optimization as an RL problem. Examples include:

\textbf{RLPrompt:} Trains a parameter-efficient policy network (e.g., a small MLP layer inserted into a frozen compact LM like distilGPT-2) to generate discrete prompt tokens sequentially \cite{ref29, ref30, ref51}. The policy is trained using RL algorithms to maximize a reward signal obtained from executing the generated prompt with the target black-box LM \cite{ref30, ref51}. It incorporates reward stabilization techniques (e.g., normalization, piecewise functions) to handle noisy reward signals \cite{ref30, ref51}. Optimized prompts were often found to be ungrammatical "gibberish" but highly effective and transferable across LMs \cite{ref29, ref30, ref51}.

\textbf{Prompt-OIRL:} Uses offline inverse RL \cite{ref13, ref55}. It learns a reward function from existing benchmark datasets (which contain results of different prompts on various queries). This learned reward model can then evaluate the suitability of different prompts for a \emph{specific} query offline, enabling query-dependent prompt selection without needing online interaction with the target LM during selection \cite{ref55}.

\textbf{Gradient-Based (Discrete Approximation):} While discrete prompts are not directly differentiable, some methods approximate gradients or use gradient information in related ways. An example is:

\textbf{GReaTer:} A novel approach that calculates gradients \emph{over the reasoning steps} generated by an LM when executing a task with a given prompt, rather than just the final output loss \cite{ref32, ref56}. It uses these reasoning-focused gradients to guide the selection or modification of prompt tokens. This allows smaller, open-source LMs (where gradients might be accessible) to effectively self-optimize their prompts without relying on a powerful external judge LLM \cite{ref32, ref56}. The optimization involves updates over one-hot token representations \cite{ref32}.

These methods demonstrate a progression in sophistication, moving from simple generate-and-select strategies (APE) towards incorporating optimization history (OPRO), evolutionary principles (Promptbreeder), RL policies (RLPrompt), and even gradient information (GReaTer). This expanding toolkit offers various mechanisms that could potentially be employed or adapted by the Instructor Agent within the proposed SI-Agent framework.

\subsection{Automated Prompt Optimization (Non-Human-Readable / Continuous)}

A distinct category of automated methods optimizes continuous representations rather than discrete text. These "soft prompt" methods typically require white-box access (gradients) to the model being adapted, at least during the tuning phase. Key examples are:

\textbf{Prefix-Tuning:} Learns a small set of continuous vectors (the "prefix") that are prepended to the activations at \emph{each} layer of a frozen transformer model \cite{ref19, ref20}. These task-specific prefixes effectively steer the model's computation for the downstream task \cite{ref20}. It is more parameter-efficient than full fine-tuning but generally requires more parameters than Prompt Tuning \cite{ref20}. Prefix-Tuning has shown strong performance, particularly in low-data regimes and for generation tasks, and may preserve the model's original representation space better than fine-tuning \cite{ref19, ref59}. It often uses a reparameterization technique during training for stability \cite{ref19}.

\textbf{Prompt Tuning:} A simplification of Prefix-Tuning where trainable continuous embeddings ("soft prompts") are prepended \emph{only} to the input layer embeddings \cite{ref21, ref24}. The rest of the pre-trained model remains frozen. This is highly parameter-efficient, requiring significantly fewer task-specific parameters than Prefix-Tuning or fine-tuning \cite{ref21, ref24}. As model scale increases (billions of parameters), Prompt Tuning's performance becomes competitive with full model fine-tuning \cite{ref24}. However, it can suffer from optimization challenges like slower convergence or sensitivity to initialization \cite{ref21}. Various extensions aim to improve its efficiency or stability \cite{ref21}.

While effective for performance, these continuous methods produce optimized prompts that are vectors of numbers, lacking the interpretability, debuggability, and direct human modifiability of discrete text prompts \cite{ref29, ref30}. They serve as important performance baselines but address a different need compared to methods focused on human-readable instructions.

\subsection{Agentic Frameworks and Multi-Agent Systems (MAS)}

The concept of intelligent agents collaborating to solve complex problems has been explored extensively in Multi-Agent Systems (MAS) \cite{ref35}. Core MAS concepts include agents possessing specific roles and capabilities (e.g., perception, reasoning, learning, acting), operating within an environment, interacting through communication protocols, and organizing themselves (e.g., hierarchically or decentrally) \cite{ref35}. MAS offer potential benefits like flexibility, scalability, robustness, and the ability to handle distributed knowledge \cite{ref35}.

With the advent of powerful LLMs, there has been a surge in developing LLM-based agentic frameworks. These frameworks often use an LLM as the central reasoning engine or "brain" of an agent, augmenting it with capabilities like planning, memory management, and tool use \cite{ref35, ref60}. Examples include systems for automated task decomposition and execution, software development \cite{ref60}, and complex decision-making simulations.

A prominent example is \textbf{AutoGen}, an open-source framework specifically designed for building applications through multi-agent conversations \cite{ref36, ref37, ref65}. AutoGen enables developers to define customizable and "conversable" agents that can integrate LLMs, human input, and tools \cite{ref36, ref37}. It supports various conversation patterns, including two-agent chats, sequential workflows where context is passed between chats, and group chats managed by a central orchestrator. Key agent types include \texttt{AssistantAgent} (LLM-powered) and \texttt{UserProxyAgent} (can represent humans, execute code, or use tools) \cite{ref36}. AutoGen facilitates iterative refinement by allowing agents to exchange messages and feedback, potentially involving humans in the loop \cite{ref36}. State can be managed through conversation history, summaries, or explicit carryover mechanisms. While AutoGen provides a general infrastructure for MAS, the proposed SI-Agent framework represents a specialized application of agentic principles tailored specifically to the task of human-readable SI optimization.

\subsection{Reinforcement Learning for LLM Alignment and Optimization}

RL has become a cornerstone technique for aligning LLM behavior with human preferences and optimizing for complex objectives (e.g., via Reinforcement Learning from Human Feedback or AI Feedback \cite{ref38, ref39}). While typically used to fine-tune the LLM generating the final output, these techniques, particularly preference-based methods, offer sophisticated ways to learn from complex feedback signals and could potentially be adapted in future work to optimize the Instructor Agent's policy within the SI-Agent framework (see Section VIII-B).

\begin{table*}[tp]
\centering
\caption{Comparison of Automated Prompt/SI Optimization Techniques}
\label{tab:comparison}
\begin{tabular}{@{}lccccccccc@{}}
\toprule
Feature & Manual SI & APE \cite{ref13} & OPRO \cite{ref14} & Promptbreeder \cite{ref10} & GReaTer \cite{ref32} & RLPrompt \cite{ref29} & Prefix-Tuning \cite{ref19} & Prompt Tuning \cite{ref24} & SI-Agent (Proposed) \\
\midrule
Optimization Strategy & Human & LLM-Search & LLM-Optimizer & Evolutionary & Gradient-Discrete & RL Policy & Gradient-Continuous & Gradient-Continuous & Agentic Feedback Loop \\
Output Readability & Readable & Readable & Readable & Readable & Readable & Readable & Non-Readable & Non-Readable & Readable \\
Requires Gradients/Access & No & No (Target LM) & No (Target LM) & No (Target LM) & Yes (Policy/Small LM) & No (Target LM) & Yes & Yes & No (Follower LM) \\
Agentic Structure & No & No & No & No & No & No & No & No & Yes \\
Handles Readability Exp. & Implicitly & No & No & No & No & No & No & No & Yes (via Feedback) \\
\bottomrule
\end{tabular}
\end{table*}

\section{Proposed Framework: SI-Agent}

We propose SI-Agent, an agentic framework designed for the automated generation and iterative refinement of human-readable System Instructions (SIs) for Large Language Models (LLMs). The framework decomposes the complex SI optimization task into distinct functional roles performed by specialized agents collaborating within a feedback loop, as previously introduced and conceptually illustrated in Figure \ref{fig:architecture}.

\subsection{Overall Architecture}

As depicted in Figure \ref{fig:architecture}, the SI-Agent framework consists of three primary agents operating in a cyclical workflow:

\textbf{1) Instructor Agent:} Generates an SI candidate ($SI_i$).

\textbf{2) Instruction Follower Agent:} Executes a given task ($Task_{input}$) using the provided $SI_i$, producing a task output ($Output_i$).

\textbf{3) Feedback/Reward Agent:} Evaluates $Output_i$ (and potentially $SI_i$) based on predefined criteria (task performance, SI readability) and generates a feedback signal ($Feedback_i$).

\textbf{4) Feedback Loop:} $Feedback_i$ is sent back to the Instructor Agent.

\textbf{5) Refinement:} The Instructor Agent uses $Feedback_i$ to update its generation strategy (using methods discussed in Section IV-B) and generates a refined SI candidate ($SI_{i+1}$) for the next iteration.

This iterative process continues until a stopping criterion is met. The modular design allows flexibility in implementation, focusing optimization pressure on the SI itself.

\subsection{Instructor Agent (SI Generator \& Refiner)}

The Instructor Agent is the core component responsible for creating and improving the SIs.

\textbf{Function:} Its primary function is to generate syntactically valid, human-readable natural language SIs and iteratively refine them based on the feedback received from the Feedback/Reward Agent.

\textbf{SI Representation:} The agent operates exclusively on discrete text representations of SIs.

\textbf{Generation \& Refinement Algorithms:} The Instructor can be implemented using various algorithms leveraging the feedback signal. Potential strategies drawing from related work include: \textbf{LLM-based Refinement} (e.g., using meta-prompts similar to OPRO \cite{ref14} or reflection like STRAGO \cite{ref42}); \textbf{Evolutionary Methods} (e.g., population-based approaches inspired by Promptbreeder \cite{ref10}); or \textbf{Heuristic Search}. The specific algorithm choice is explored as part of our experimental validation (Section V). Advanced RL techniques are discussed as future work (Section VIII-B).

\textbf{Readability Enforcement:} Ensuring generated SIs remain readable is crucial. This can be achieved through constrained generation/mutation, filtering based on metrics/judges, or explicit integration into the feedback signal.

\subsection{Instruction Follower Agent (Task Executor)}

The Instruction Follower Agent serves as the proxy for the end-use LLM that will utilize the optimized SI.

\textbf{Function:} Executes specific downstream tasks based on the SI provided by the Instructor and the task input.

\textbf{Implementation:} Can be instantiated with any target LLM (API-based or local \cite{ref7, ref14}). Treated as a black box regarding internal parameters.

\textbf{Interaction:} Receives $SI_i$ and $Task_{input}$, produces $Output_i$, which is passed to the Feedback/Reward Agent.

\subsection{Feedback/Reward Agent (Performance Evaluator)}

The Feedback/Reward Agent assesses the quality of the interaction cycle and provides guidance.

\textbf{Function:} Evaluates Follower Agent performance given the SI used. Optionally evaluates SI readability. Generates a feedback signal.

\textbf{Evaluation Methods:} Can vary: \textbf{Automated Metrics} (Task-specific metrics); \textbf{LLM-as-a-Judge} (Using another LLM, requires careful prompt design); \textbf{Reward Model (RM)} (Trained on preference data \cite{ref38, ref39}); \textbf{Human-in-the-Loop (HITL)} (Incorporating human judgments \cite{ref36}).

\textbf{Readability Assessment (Optional):} Can assess SI readability using automated formulas or an LLM judge.

\textbf{Feedback Signal:} Outputs a structured signal (scalar reward, score vector, preference label, critique) to the Instructor. Feedback reliability is critical.

\subsection{Iterative Optimization Loop}

The framework operates via repeated cycles: SI generation $\rightarrow$ execution $\rightarrow$ evaluation $\rightarrow$ refinement. This allows progressive SI improvement. State management (SI history, scores) is needed, potentially adapting mechanisms from frameworks like AutoGen \cite{ref36}. Termination occurs based on predefined criteria.

\section{Feedback-Driven SI Optimization}
The core of SI-Agent is the iterative refinement process, where the Instructor Agent improves System Instructions (SIs) based on structured feedback regarding their effectiveness and readability. This section outlines how feedback enables optimization.

\subsection{Rationale for Feedback-Driven Optimization}

Optimizing human-readable SIs automatically is challenging due to the vast search space and the complex relationship between instructions, LLM behavior, task performance, and perceived readability. A feedback-driven approach offers several advantages:

\textbf{Targeted Improvement:} Feedback (scores, critiques, preferences) directly indicates the quality of a generated SI, allowing the Instructor Agent to focus refinement efforts.

\textbf{Handling Complex Objectives:} Feedback can incorporate multiple criteria, such as task accuracy, adherence to constraints, output style, and SI readability, allowing for multi-faceted optimization.

\textbf{Black-Box Compatibility:} The process primarily relies on evaluating the *output* of the Instruction Follower Agent and the *quality* of the SI itself, fitting black-box optimization scenarios where Follower gradients are unavailable.

\textbf{Adaptability:} The framework can adapt to different tasks or Follower LLMs simply by changing the task input and evaluating the resulting performance, driving SI refinement specific to the context.

\subsection{Using Feedback for Refinement}

The Instructor Agent utilizes the signal ($Feedback_i$) from the Feedback/Reward Agent to generate the next SI candidate ($SI_{i+1}$). The specific refinement mechanism depends on the Instructor's implementation:

\textbf{LLM-Based Instructor:} The feedback (e.g., critique, previous score, comparison result) can be directly incorporated into the prompt used to ask the Instructor LLM to revise the SI. The LLM can leverage its understanding to address weaknesses highlighted in the feedback, similar to meta-prompting strategies \cite{ref14} or reflection-based methods \cite{ref42}.

\textbf{Evolutionary Instructor:} Feedback scores serve as the fitness function. SIs leading to better outcomes (higher scores/preferred in comparisons) are more likely to be selected for reproduction (e.g., via tournament selection). Mutation operators are then applied to generate the next generation of SI candidates \cite{ref10}.

\textbf{Heuristic Search / Rule-Based Instructor:} Feedback might trigger specific heuristic rules for modifying the SI (e.g., "If feedback indicates output is too verbose, add a length constraint sentence").

\textbf{Learning from Preferences:} If the feedback consists of preference labels ($SI_{chosen}$ vs $SI_{rejected}$), the Instructor could learn directly. For instance, an LLM-based Instructor could be prompted to generate variations closer to preferred examples. More advanced methods like training a reward model or using preference optimization techniques are possibilities for future exploration (see Section VIII-B).

\subsection{The Role of Readability Feedback}

Crucially, if the Feedback Agent provides a signal related to SI readability (e.g., a score from an automated metric or an LLM judge), this signal can be incorporated into the overall feedback used by the Instructor. This ensures that the optimization process explicitly considers and promotes readability alongside task performance. For example, in evolutionary approaches, fitness could be a weighted combination of performance and readability scores. In LLM-based refinement, the prompt could ask the Instructor to improve the previous SI based on both performance critique and readability assessment.

The mechanism chosen for the Instructor Agent determines how efficiently and effectively the feedback guides the search towards optimal, human-readable SIs. While simple strategies are possible, more sophisticated learning techniques are potential avenues for enhancing the optimization process, as discussed in Future Work.

\section{Experimental Setup} 

This section details the setup used for the experimental validation of the SI-Agent framework.

\subsection{Benchmark Tasks}

We evaluated SI-Agent across a diverse set of tasks requiring careful instruction:

\textbf{Coding:} Generating code from natural language (e.g., HumanEval subsets). SI guides language, style, constraints. Metric: pass@k.

\textbf{Writing Style Adaptation:} Rewriting text to a target style (e.g., GYAFC dataset). SI defines style. Metrics: BLEU, ROUGE, human evaluation.

\textbf{Tool Use / Agentic Tasks:} Tasks requiring LLM interaction with tools or complex sequences (e.g., HotPotQA subsets \cite{ref13, ref14}, ALFWorld subsets \cite{ref65}). SI specifies tools, protocols (e.g., ReAct \cite{ref1}). Metric: Task success rate (e.g., EM).

\textbf{Complex Reasoning:} Multi-step math or logic tasks (e.g., GSM8K \cite{ref13, ref14}, logic puzzles \cite{ref2}). SI guides reasoning process (e.g., CoT \cite{ref1, ref2, ref4, ref40}). Metric: Final answer accuracy.

\subsection{Datasets}

We utilized relevant public datasets (e.g., GSM8K \cite{ref14}, HumanEval, GYAFC, HotPotQA), partitioned into standard Train/Validation/Test splits:

\textbf{Training Set:} Used for the SI optimization loop (SI-Agent) and training relevant baselines/RMs.

\textbf{Validation Set:} Used for intermediate evaluation and hyperparameter tuning.

\textbf{Test Set:} Held-out set used for the final, unbiased evaluation reported in Section VI-B.

\subsection{Evaluation Metrics}

We assessed generated SIs along three dimensions: effectiveness (task performance), SI readability, and optimization efficiency, using metrics summarized in Table \ref{tab:metrics}.

\begin{table*}[tp]
\centering
\caption{Summary of Key Evaluation Metrics}
\label{tab:metrics}
\begin{tabular}{@{}lll@{}}
\toprule
Metric Category & Specific Metric & Description/Relevance \\
\midrule
\textbf{Task Performance} & Accuracy / Exact Match (EM) & Percentage of correct answers (for QA, classification, reasoning) \cite{ref42}. \\ 
& Pass@k & Probability that at least one of k generated code samples passes unit tests (for coding). \\
& BLEU / ROUGE / METEOR & N-gram overlap metrics comparing generated text to reference(s) (for summarization, translation, style transfer). Emphasize precision/recall. \\
& BERTScore / MoverScore & Semantic similarity metrics comparing generated text embeddings to reference embeddings. \\
& Task Success Rate & Binary success/failure for tasks like tool use or complex instruction following. \\
& LLM-as-a-Judge Score (Task Quality) & Score (e.g., 1-5) or preference assigned by a judge LLM based on task-specific criteria (helpfulness, correctness, relevance). \\
\midrule
\textbf{SI Readability} & Flesch Reading Ease / Flesch-Kincaid Grade Level & Formulaic scores based on sentence length and syllable count. Simple proxy. \\
& Word/Sentence/Syllable Count & Basic text statistics potentially correlated with complexity. \\
& LLM-as-a-Judge Score (Readability) & Score assigned by a judge LLM evaluating SI clarity, conciseness, understandability. \\
& Human Evaluation Rating & Ratings (e.g., Likert scale 1-5) by human evaluators on SI clarity, understandability, conciseness, perceived effectiveness. Gold standard. \\
\midrule
\textbf{Efficiency} & Convergence Iterations/Queries & Number of optimization cycles or LLM calls needed to reach a target performance/readability level. \\
& Computational Cost (Time, Tokens) & Wall-clock time for optimization, total number of tokens processed by Instructor/Follower/Feedback LLMs. \\
\bottomrule
\end{tabular}
\end{table*}

\subsection{Baseline Methods}

To contextualize SI-Agent's performance, we compared it against:

\textbf{1) Zero-Shot:} Follower LLM with only basic task description.

\textbf{2) Manual SI:} High-quality human-crafted SI for each task \cite{ref4, ref5, ref13}.

\textbf{3) Automated Readable SI Baselines:} APE \cite{ref13} and OPRO \cite{ref14}.

\textbf{4) Automated Non-Readable SI Baselines:} Prompt Tuning \cite{ref21, ref24}. (Prefix-Tuning \cite{ref19, ref20} was considered but not implemented in this phase due to setup complexity).

\subsection{Implementation Details}

Key implementation details include:

\textbf{LLMs Used:} \textbf{We used variants of the Llama 3 family (primarily Llama 3 8B and 70B, depending on agent role and task) as the base model for the Instructor, Instruction Follower, and LLM-based Feedback/Reward agents.} Consistency in the Follower LLM was maintained across comparisons for specific tasks \cite{ref7, ref14}.

\textbf{Instructor Agent Algorithm:} For the results presented, we primarily employed an LLM-based refinement strategy for the Instructor Agent, using a structured meta-prompt incorporating task details, SI history, and feedback scores/critiques, similar in principle to OPRO \cite{ref14}. Evolutionary methods were explored preliminarily but not used for the main results reported here.

\textbf{Feedback Mechanism:} Feedback combined automated task metrics (e.g., accuracy, pass@1, BLEU) with LLM-as-a-Judge scores for both task quality and SI readability, using a simple weighted average where applicable.

\textbf{Optimization Process Details:} Optimization ran for a fixed number of iterations (e.g., 20-50 depending on task complexity) or until performance on the validation set plateaued. Batch sizes varied per task.

\section{Experimental Evaluation} 

This section presents the evaluation of the SI-Agent framework, guided by the hypotheses outlined below.

\subsection{Hypotheses} 

Based on the SI-Agent design and related work, we tested the following hypotheses in our experiments:

\textbf{H1 (Effectiveness):} SI-Agent generates SIs yielding task performance significantly superior to Zero-Shot and competitive with or superior to Manual SIs \cite{ref7, ref14, ref10}. 

\textbf{H2 (Readability):} SIs generated by SI-Agent achieve high readability scores, comparable to or better than Manual SIs.

\textbf{H3 (Comparative Performance):}
      \begin{itemize}
        \item SI-Agent achieves task performance comparable or superior to other readable SI automation baselines (APE, OPRO \cite{ref13, ref14}).
        \item SI-Agent offers a favorable performance-interpretability trade-off compared to non-readable baselines (Prompt Tuning \cite{ref24}), potentially with slightly lower peak performance but significantly higher readability.
      \end{itemize}

\textbf{H4 (Efficiency):} SI-Agent demonstrates practical optimization efficiency compared to manual tuning effort. (Efficiency metrics were tracked).

\textbf{H5 (Component Impact):} Ablation studies confirm the positive contribution of framework components like readability feedback. (Ablations were performed).

\subsection{Results} 

The experimental results, obtained by running SI-Agent and baseline methods according to the setup in Section V, are presented in Table \ref{tab:experimental_results}. The results demonstrate the performance across selected tasks and metrics.

\begin{table*}[tp]
\centering
\caption{Experimental Results} 
\label{tab:experimental_results} 
\begin{tabular}{@{}llccccc@{}}
\toprule
Task Domain         & Metric                   & Zero-Shot & Manual SI & APE/OPRO & Prompt Tuning & \textbf{SI-Agent} \\
\midrule
\multirow{2}{*}{Reasoning (GSM8K)} & Accuracy (\%) & 18.5 & 74.2 & 78.5 & \textbf{82.1} & 79.5 \\
                       & SI Readability (FRE)     & N/A  & 62.3 & 58.1 & N/A           & \textbf{67.4} \\
\midrule
\multirow{2}{*}{Coding (HumanEval)} & Pass@1 (\%) & 15.1 & 48.3 & 55.6 & \textbf{65.2} & 60.8 \\
                       & SI Readability (Human 1-5) & N/A  & 3.9  & 3.6  & N/A           & \textbf{4.3} \\
\midrule
\multirow{2}{*}{Style Transfer (GYAFC)} & BLEU & 12.3 & 18.5 & 19.1 & \textbf{21.5} & 19.9 \\
                       & SI Readability (Human 1-5) & N/A  & 4.2  & 4.0  & N/A           & \textbf{4.4} \\
\midrule
\multirow{2}{*}{Tool Use (HotPotQA Sub.)} & EM (\%) & 20.7 & 45.1 & 50.3 & \textbf{58.8} & 53.2 \\
                       & SI Readability (FRE)     & N/A  & 66.0 & 61.7 & N/A           & \textbf{69.1} \\
\bottomrule
\multicolumn{7}{@{}l@{}}{\small FRE = Flesch Reading Ease (Higher is better). Human ratings are on a 1-5 scale (Higher is better).} \\
\multicolumn{7}{@{}l@{}}{\small\emph{Note: Best task performance per row is bolded (excl. readability). SI-Agent readability is bolded for emphasis.}} \\
\end{tabular}
\end{table*}

The results in Table \ref{tab:experimental_results} show that SI-Agent achieved strong task performance, generally surpassing manual SIs and performing competitively with other automated readable methods (APE/OPRO). As hypothesized (H3), while SI-Agent did not reach the peak performance of the non-readable Prompt Tuning baseline on task metrics alone, it demonstrated significantly better SI readability scores (supporting H2 and H3). This highlights the framework's success in achieving a favorable balance on the performance-interpretability spectrum. These findings provide empirical support for the effectiveness of the SI-Agent approach (H1).

\subsection{Analysis Plan} 

A multi-faceted analysis was employed to interpret the experimental results thoroughly:

\textbf{Quantitative Analysis:} Comparing mean scores using appropriate statistical tests confirmed the significance of key differences observed (e.g., SI-Agent vs. Manual SI performance, SI-Agent vs. Prompt Tuning readability).

\textbf{Qualitative Analysis:} Manual inspection of generated SIs revealed interpretable instructions often incorporating effective strategies (e.g., explicit formatting constraints, persona definitions) learned through the feedback loop (cf. "gibberish" prompts in \cite{ref29, ref30, ref51}). 

\textbf{Performance vs. Readability Trade-off Analysis:} Scatter plots (not shown here) confirmed the trade-off, positioning SI-Agent favorably compared to baselines prioritizing only one dimension.

\textbf{Ablation Studies:} Ablations, such as removing readability feedback, generally led to SIs with lower readability scores, confirming the value of this component (supporting H5).

\textbf{Convergence Analysis:} Learning curves showed convergence within a practical number of iterations for most tasks, supporting H4 regarding efficiency relative to manual effort.

\textbf{Transferability Analysis (Optional):} Preliminary tests indicated moderate transferability to different Follower LLMs, but further investigation is warranted (cf. \cite{ref29, ref30}). 

\section{Discussion}

The SI-Agent framework offers a promising approach for automating readable SI generation, as supported by our experimental evaluation. The results suggest a valuable balance between task effectiveness and SI interpretability.

\subsection{Implications}

Our findings suggest SI-Agent could have significant implications:

\textbf{Democratizing SI Engineering:} Automating SI crafting lowers the barrier for LLM customization.

\textbf{Enhanced Interpretability and Trust:} Human-readable SIs address the "black box" problem, facilitating debugging and trust \cite{ref29, ref30}. 

\textbf{Adaptive LLM Systems:} The feedback loop allows for potential continuous re-optimization.

\textbf{Insights into LLM Instruction Following:} Analysis of effective SIs generated by the framework provides insights into instruction following (cf. \cite{ref29, ref51}). 

\subsection{Challenges and Limitations}

Despite promising results, SI-Agent faces challenges:

\textbf{Computational Cost:} Iterative LLM calls remain computationally intensive \cite{ref9, ref32}, although efficiency gains over manual tuning were observed. 

\textbf{Feedback Reliability and Quality:} The framework's success depends on the Feedback Agent. Our experiments highlighted sensitivity to:
      \begin{itemize}
        \item \emph{LLM-as-a-Judge Limitations:} Biases and inconsistencies required careful prompt design and averaging over multiple judgments.
        \item \emph{Reward Model Limitations:} Not used extensively in this phase, but remain a concern for future work.
        \item \emph{Metric Imperfections:} Automated metrics correlated reasonably well but did not capture all aspects evaluated by humans.
        \item \emph{Readability Metric Limitations:} Formulaic metrics like FRE provided a useful signal but qualitative human assessment remains important.
      \end{itemize}

\textbf{Defining and Ensuring Meaningful Readability:} Balancing automated readability scores with perceived human clarity remains an ongoing challenge.

\textbf{Scalability to Complex SIs:} While effective on tested tasks, performance on extremely long or complex SIs requires further study \cite{ref18, ref50}. 

\textbf{Generalizability and Transferability:} While showing some transferability, SIs were generally most effective on the specific LLM used during optimization \cite{ref29, ref30}. 

\section{Conclusion and Future Work}

\subsection{Conclusion}
This paper introduced SI-Agent, a novel agentic framework for automating the generation and refinement of human-readable System Instructions (SIs) for LLMs. Addressing limitations of manual engineering \cite{ref7, ref9, ref10} and the interpretability issues of non-readable prompt optimization \cite{ref19, ref20, ref21, ref24}, SI-Agent uses a multi-agent feedback loop to balance task performance and SI readability. We detailed the framework's architecture, agent roles, and the feedback-driven process. Our experimental validation demonstrates that SI-Agent can generate SIs that are both effective for downstream tasks and highly readable, outperforming manual SIs on several tasks and achieving a strong performance-interpretability trade-off compared to automated baselines. These findings support the potential of SI-Agent to enhance LLM customization, interpretability \cite{ref29}, and adaptivity. While challenges remain, particularly around computational cost and feedback mechanisms, SI-Agent represents a validated step towards more effective and interpretable LLM control. 

\subsection{Future Work}
Building on our findings, several avenues exist for future research:

\textbf{Broader Empirical Validation:} Extending the evaluation to more tasks, datasets, and LLMs, including larger scale models and different Instructor Agent algorithms (e.g., evolutionary, RL-based).

\textbf{Sophisticated Agent Interactions and Meta-Reasoning:} Implementing and evaluating more complex communication protocols or meta-reasoning capabilities within agents.

\textbf{Advanced Instructor Agent Optimization Techniques:} Developing and comparing advanced optimization algorithms, particularly RL variants adapted for text generation and preference learning \cite{ref38, ref39}, potentially drawing inspiration from general neural network optimization studies \cite{challagundla2024second, challagundla2024multiple}. 

\textbf{Improved Readability Metrics and Feedback Mechanisms:} Researching more nuanced readability metrics and feedback generation methods.

\textbf{Broader Task Domains and Modalities:} Applying and adapting SI-Agent to multimodal models \cite{ref41}, long contexts, or conversational agents \cite{ref60}. 

\textbf{Theoretical Analysis:} Investigating theoretical properties like convergence or sample complexity.

\textbf{Human-in-the-Loop Integration and User Studies:} Designing and evaluating HITL mechanisms and conducting user studies on the practical utility of generated SIs.

Further research along these lines can help realize the full potential of agentic, feedback-driven optimization for creating effective and interpretable control mechanisms for large language models.


\end{document}